# Robust Privacy amidst Innovation with Large Language Models through a Critical Assessment of the Risks

Yao-Shun Chuang[1], Atiquer Rahman Sarkar[2], Yu-Chun Hsu[1], Noman Mohammed[2], and Xiaoqian Jiang[1]

[1] McWilliams School of Biomedical Informatics, The University of Texas Health Science Center at Houston, Houston, Texas, USA
[2] Dept. of Computer Science, University of Manitoba, Winnipeg, R3T 5V6, Canada

**Abstract**
*Objective*

This study evaluates the integration of Electronic Health Records (EHRs) and Natural Language Processing (NLP) with large language models (LLMs) to enhance healthcare data management and patient care, focusing on using advanced language models to create secure, HIPAA-compliant synthetic patient notes for global biomedical research.

*Materials and Methods*

The study used de-identified and re-identified versions of the MIMIC III dataset with GPT-3.5, GPT-4, and Mistral 7B to generate synthetic clinical notes. Text generation employed templates and keyword extraction for contextually relevant notes, with one-shot generation for comparison. Privacy was assessed by analyzing PHI occurrence and co-occurrence, while utility was evaluated by training an ICD-9 coder using synthetic notes. Text quality was measured using ROUGE and cosine similarity metrics to compare synthetic notes with source notes for semantic similarity.

*Results and Discussion*

Analysis of PHI occurrence and text utility via the ICD-9 coding task showed that the keyword-based method had low risk and good performance. One-shot generation showed the highest PHI exposure and PHI co-occurrence, especially in geographic location and date categories. The Normalized One-shot method achieved the highest classification accuracy. Privacy analysis revealed a critical balance between data utility and privacy protection, influencing future data use and sharing. Re-identified data consistently outperformed de-identified data.

*Conclusion*

This study shows that keyword-based methods can create synthetic clinical notes that protect privacy while retaining data usability, potentially improving clinical data sharing. The use of dummy PHIs to counter privacy attacks may offer better utility and privacy than traditional de-identification.

**Introduction**

Electronic health records (EHRs) are vital digital tools in healthcare, storing detailed patient information such as demographics, medical history, and lab results[1]. Efficient data management, enhanced patient care, and research and analysis are pivotal aspects brought forward by integrating EHRs. EHRs are instrumental in streamlining patient information, significantly improving healthcare delivery by making it more efficient and effective[2,3]. Furthermore, previous studies emphasized that EHRs improve patient care by providing immediate access to histories and data for swift decision-making and enhancing medical research and public health through their ability to facilitate detailed analysis of health trends and outcomes[2,4]. Together, these aspects underscore the transformative impact of EHRs on healthcare, from improving individual patient care to advancing medical research and public health initiatives.

Natural language processing (NLP) technologies harness the extensive unstructured data within EHRs. NLP helps interpret and extract meaningful information from unstructured data, converting it into a structured format[5]. This transformation enhances patient care and healthcare operations by enabling the efficient use and analysis of data[6]. With structured data, researchers and healthcare professionals can conduct secondary analyses of EHR information[7,8]. The benefits include informing large-scale health systems choices, helping in making point-of-care clinical decisions, and improving patient safety and health[9–11]. These are critically significant in informing public health policy and advancing research to provide retrospective analysis, investigation into long-term effects, and an overview of trends.

The widespread adoption of EHRs ushers a new era of healthcare research by providing researchers access to invaluable data. However, this data sharing introduces significant privacy concerns, particularly around protected health information (PHI). The Health Insurance Portability and Accountability Act (HIPAA) plays a crucial role in safeguarding medical information in the U.S. NLP techniques are used to de-identify PHI to facilitate data use for research and other secondary purposes while preserving privacy; however, despite advancements in enhancing data privacy, these NLP frameworks and models have not yet fully complied with HIPAA's stringent standards[12–14]. The challenge of creating publicly accessible datasets for research without compromising privacy remains unresolved. This scarcity of data impedes the development of NLP models by limiting their generalizability and shareability[11,15,16]. These significant gaps underscore the need for continued research to improve these models, ensuring they meet HIPAA standards and adhere to ethical guidelines for using health data within the healthcare industry.

This study presents a novel method that enhances data privacy and interoperability in biomedical research by using advanced large language models (LLMs) to generate private, high-quality synthetic notes. This method surpasses existing data sharing and privacy techniques, facilitating global collaboration while ensuring that synthetic notes are anonymized and maintaining actual patient data's usability. Additionally, it improves trust and transparency in patient data usage, encouraging broader participation in clinical research by ensuring data security and privacy. This approach not only sets new ethical standards for AI in healthcare but also represents a paradigm shift in maintaining patient privacy in the digital health era.

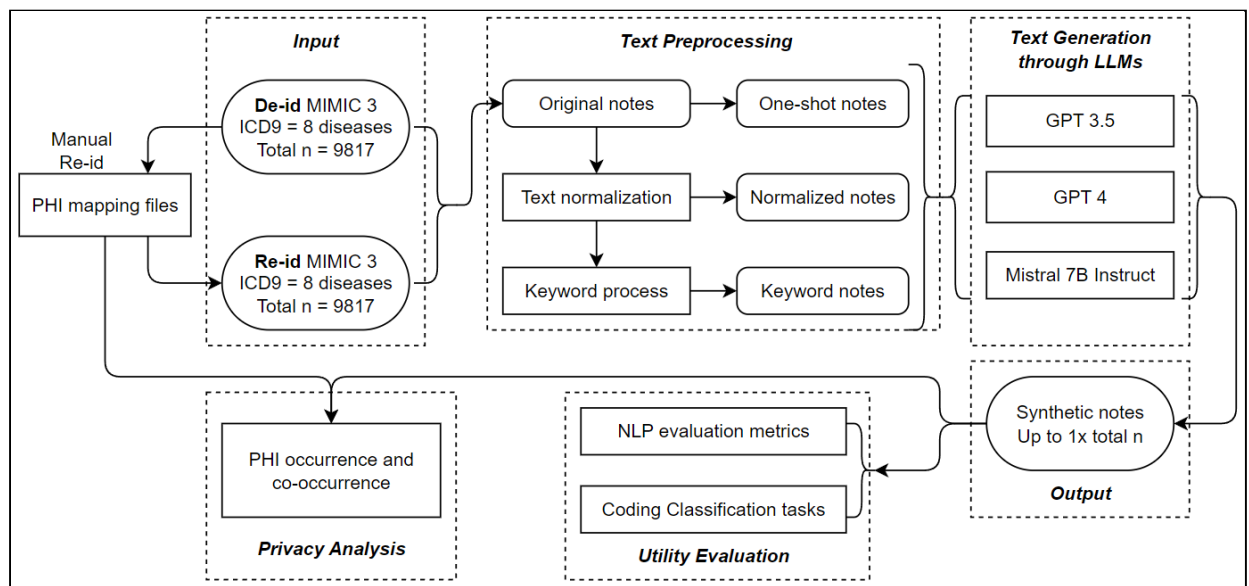

**Figure 1**. A flowchart that provides an overview of our study.

## Methods
### Dataset
In this study, a subset of the MIMIC III database was utilized, comprising 9,817 discharge summaries. Initially, this dataset featured a multi-ICD-9 labeling system, encoding multiple diagnoses for each patient. Following the methodology outlined in the study, the dataset was filtered and simplified to single-label entries, focusing exclusively on the primary diagnosis[17]. This process involved selecting only the main diagnosis for each patient and narrowing the scope to eight major diseases. These diseases, defined by their respective ICD-9 codes, include hypertension, congestive heart failure, atrial fibrillation, kidney failure, type II diabetes, respiratory failure, and urinary tract infection.

**Models**

Three advanced language models, GPT-3.5-0301 (GPT3.5), GPT-4-0613 (GPT4), and Mistral7B-Instruct (Mistral7B), were utilized in this study for generating synthetic clinical notes. Each model was chosen based on its specific capabilities to produce coherent and contextually relevant text. GPT-3.5 and GPT-4, iterations of the Generative Pre-trained Transformer models developed by OpenAI, are among the most popular LLMs. These models were selected not only for their widespread accessibility and frequent use in daily life but also due to a lack of detailed privacy analysis beyond OpenAI's announcements. The findings from this study aim to establish a benchmark for understanding privacy risks associated with these models. Additionally, Mistral7B is a robust instruction-based LLM with only 7 billion parameters[18]. The study showed that it can achieve performance comparable to more complex models in text generation tasks by adhering strictly to user instructions[19]. Mistral7B may offer superior synthetic notes with excellent usability because it can produce high-quality, human-like synthetic text. The GPT series models, accessed via the OpenAI API, were deployed on the HIPAA-compliant Azure OpenAI platform provided by UTHealth to comply with the data usage agreement requirements of MIMIC-III.

**Text preprocessing and generation**

Data re-identification (RE-ID) was performed on the MIMIC III database. The original data had been de-identified (DE-ID), with PHI replaced by 27 unique PHI tags, each enclosed within brackets to indicate redaction. These tags are classified into eight categories of PHI as defined by HIPAA. By employing a process of PHI mapping, the original notes are transformed into a dataset containing PHI. The flowchart of this study was in Figure 1. For detailed mappings and specific configurations related to each PHI category, refer to Supplementary Appendix 1, Table 1. It was noteworthy that for certain categories, such as "*university*" and *"hospital"* from MIMIC III, generating mock data posed a challenge. Consequently, these were replaced with a prefixed category tag that included the note's ID. Specifically, the PHI tag "*num*", found in the MIMIC III notes, posed challenges in understanding its semantic implications due to its ambiguous context, which made it difficult to accurately interpret whether it referred to numerical values, medical measurements, or other quantitative data. Despite being substituted with a random number to preserve the formatting, this particular tag was omitted from the PHI occurrence testing.

Three types of templates—original notes (One-shot), normalized notes (Normalized One-shot), and keyword notes (Keyword) —were employed in the text generation process, following the established sequence of data preprocessing. Consequently, one-shot prompt generation was conducted using these templates to feed into LLMs for the production of synthetic notes. The original notes, which were raw clinical notes, were directly used in the prompts. The examples of the prompts utilized are presented in the Figure 1 and Figure 2 in the Supplementary Appendix 1. During the normalization of these clinical notes, the text was refined by removing all numbers and symbols, preserving only the alphabetic words. This critical step eliminated numeric data, punctuation, and special characters, resulting in normalized notes that represented a simplified version of the text, focusing on core linguistic elements.

The NLP technique, keyword extraction process, was employed to analyze the normalized notes with the goal of extracting key phrases using the unsupervised KP-miner and YAKE algorithms[20–22]. KP-Miner utilizes a weighting mechanism that incorporates frequency and positional information to identify multi-token keywords. In contrast, YAKE applies a range of statistical measures, including positional data, frequency, and contextual information, to rank keywords. Campos et al. (2020) conducted a comprehensive comparison of various unsupervised keyword extraction methods, including RAKE and graph-based approaches such as TextRank. Their study demonstrated that YAKE outperformed these alternative methods in most cases across a wide range of twenty public datasets[22]. This consistent performance across multiple datasets was a crucial factor in our decision to use YAKE, as it provided superior results in extracting relevant keywords compared to RAKE, TextRank, and other similar approaches. In addition, studies have demonstrated KP-Miner's superior performance over other techniques and optimal results when combined with YAKE for indexing news articles[23,24]. After generating the key phrases, the "fuzzywuzzy" package was applied to remove duplicates by utilizing the Levenshtein distance to compare and withdraw similar key phrases, thereby enhancing the accuracy and relevance of the extracted phrases[25]. The output from these NLP techniques produced a cloze test-like context, which was then used to prompt LLMs to fill in the blanks, thus generating synthetic notes. This approach ensured that the emphasis remained exclusively on the important textual data, which is crucial for an accurate interpretation of the patient's medical condition as described in the notes.

**Privacy Analysis and Text Quality Evaluation**
Three distinct analyses were conducted in this study: an examination of the occurrence and co-occurrence of PHI to assess privacy risks, a text utility analysis involving a classification task, and a comparison of text similarity between real and synthetic notes. Initially, for the assessment of PHI occurrence, the synthetic notes were scanned using known PHI categories to calculate the occurrence rates of 18 different PHI types. Any PHI found in the synthetic notes, the corresponding categories only counted once for a note, and the final number was divided by 9,817 to derive the average number of notes suffering a leakage. Additionally, co-occurrence checks were performed to identify any links within the notes by observing the exposure of various PHI types.

Secondly, for the utility assessment, the ICD-9 coding classification was utilized via the LAAT package[26]. The LAAT package was selected for its implementation of a label attention model specifically designed for automatic ICD coding. This model efficiently handles varying text lengths and the interdependencies among fragments linked to ICD codes, making it well-suited for the diverse synthetic notes generated by our methodology. The package also features a hierarchical joint learning mechanism that extends the label attention model to address data imbalance issues caused by the rarity of many ICD codes. In the dataset, clinical notes for type 2 diabetes mellitus and urinary tract infection are notably fewer (410 and 800 notes, respectively) compared to approximately 3,500 notes for hypertension, out of a total of 8 ICD-9 conditions. The hierarchical approach leverages relationships among ICD codes, enhancing performance, especially for less frequent codes. These capabilities made the LAAT package the optimal choice for our study. Notably, a 10-fold cross-validation was performed to ensure the accuracy and reliability of the coding classification results. The test set of synthetic notes were mapped back to the corresponding source notes, serving as the final test set. Benchmarks for comparison included both re-identified and de-identified source notes. In the evaluation for the classification, micro- and marco-averages were provided in the evaluation results. Micro-averaging pools all decisions to compute a single effectiveness measure, emphasizing overall accuracy and the impact of larger categories. Macro-averaging calculates measures separately for each label and averages them, ensuring balanced performance across all categories regardless of their size. Each method highlighted different aspects of model performance.

Finally, traditional NLP analyses, specifically ROUGE (Recall-Oriented Understudy for Gisting Evaluation) and Cosine Similarity (COS-SIM), were employed to assess the similarity between the real and synthetic notes. ROUGE measures the overlap of n-grams between the generated text and reference texts, which offers insights into the precision and recall of the text generated by the model. For simplification in the final report, the F1-score was calculated. On the other hand, Cosine Similarity quantifies the cosine of the angle between two vectors representing the TF-IDF (Term Frequency-Inverse Document Frequency) weighted texts. This metric gauges the semantic similarity between the generated and reference texts, reflecting the model's ability to preserve the semantic content of the text. Thus, the application of ROUGE and Cosine Similarity provides a rigorous evaluation of both the quality and semantic accuracy of synthetic texts when compared to their original counterparts.

**Results**
The objective was to generate a total of 9,817 synthetic notes. Due to content filtering (e.g., perceived as potentially harmful) by the Azure OpenAI service, the notes produced by GPT-3.5 and GPT-4 experienced deficits, with up to 13 prompts for notes denied. This issue was more frequently observed in the Normalized One-shot. In contrast, text generation using the local model, Mistral7B, did not encounter this limitation. Additionally, the mean number of extracted keyword tokens across ICD9 diseases, as well as the accumulated input tokens used for each text generation method in both re-identified and de-identified data, can be found in Figures 3 and 4, respectively, in Supplementary Appendix 1. The corresponding costs associated with using GPT models are presented in Table 2 of Supplementary Appendix 1.

Using the PHI mapping file to repopulate the de-identified source notes, Table 1 presents the prevalence of PHI occurrences in synthetic notes produced by various GPT models and the Mistral7B model across different generation methodologies. All PHI categories except SSNs were identified in the synthetic notes. One-shot generation exhibited the highest risk of PHI exposure, whereas Keyword generation showed the lowest, limited primarily to geographic location PHI. In the One-shot approach, geographic location, date, and unique identifying numbers were the top three PHI categories with the highest occurrence rates in that order. Notably, the geographic location PHI had the highest prevalence across all models and generation methods. For instance, in the One-shot generation using the Mistral7B model, geographic and date PHI occurrences were notably high, at 85.35% and 87.45%, respectively, with unique identifying numbers documented at 10.40%. In contrast, during Keyword generation using the Mistral7B model, the occurrence of geographic PHI was 29.73%, while the occurrence of date PHI was 0.02%. Furthermore, the GPT3.5 and GPT4 models generally demonstrated lower PHI occurrence rates across most categories when compared to Mistral7B. For example, the Normalized One-shot generation with the GPT4 model recorded only 1.66% in names, 0.25% in unique identifying numbers, 0.11% in telephone numbers, and 0.01% in device identifiers.

In the analysis of PHI co-occurrence, the One-shot generation method exhibited the highest incidence of cases in Table 2. The GPT4 and Mistral7B models each recorded approximately 500 instances where two unique types of PHIs co-occurred in the same notes, and around ten notes contained three types of PHIs. Conversely, the GPT3.5 model displayed 270 cases with two types of PHIs and seven instances with three types. Notably, GPT3.5 and GPT4 reported one instance each where four unique types of PHIs occurred in the same note, indicating a higher risk of privacy exposure. All of these PHI co-occurrences involved the PHI category "Names." Further investigation revealed that these occurrences were primarily due to "Name initials," constituting a significant portion of the "Names" PHI across the 18 HIPAA categories in all three models. Of these, only the Mistral7B model included 64 "Full names" instances out of 144 "Name" PHI in the Normalized One-shot generation. Additionally, the "Date" category was involved in all PHI co-occurrence cases in the One-shot generation.

**Table 1.** PHI occurrence in the synthetic notes for different generation approaches when the generation is based on re-identified notes

|  | **Keyword** | | | **Normalized one-shot** | | | **One-shot** | | |
|---|---|---|---|---|---|---|---|---|---|
|  | GPT3.5 | GPT4 | Mistral7B | GPT3.5 | GPT4 | Mistral7B | GPT3.5 | GPT4 | Mistral7B |
| **Names** |  |  |  | 3.51% | 1.66% | 1.47% | 2.93% | 5.23% | 4.95% |
| **Geographic** | 23.21% | 29.98% | 29.73% | 58.60% | 57.04% | 50.20% | 32.18% | 70.31% | 85.35% |
| **All date** |  |  | 0.02% | 0.03% |  | 0.02% | 39.58% | 66.13% | 87.45% |
| **Telephone** |  |  |  | 0.24% | 0.11% | 0.07% | 0.14% | 0.21% | 0.53% |
| **Device identifiers** |  |  |  | 0.06% | 0.01% | 0.14% | 0.13% | 0.20% | 0.25% |
| **Medical record numbers** |  |  |  | 0.01% |  | 0.01% | 0.03% | 0.02% | 0.12% |
| **Unique identifying number** |  |  |  | 0.69% | 0.25% | 1.35% | 9.56% | 25.42% | 10.40% |

**Table 2.** The Prevalence of PHI Co-occurrence

| Models | | *2 PHIs* | | *3 PHIs* | | *4 PHIs* | |
|---|---|---|---|---|---|---|---|
| *Normalized One-shot* | GPT3.5 | 4 | (0.04%) |  |  |  |  |
|  | GPT4 | 2 | (0.02%) |  |  |  |  |
|  | Mistral7B |  |  |  |  |  |  |
| *One-shot* | GPT3.5 | 270 | (2.75%) | 7 | (0.07%) | 1 | (0.01%) |
|  | GPT4 | 490 | (4.99%) | 9 | (0.09%) | 1 | (0.01%) |
|  | Mistral7B | 567 | (5.78%) | 13 | (0.13%) |  |  |

The evaluation of precision for ICD-9 coding is shown in Figure 2. The detailed values can be found in Table 3 and Table 4 of Supplementary Appendix 1. The micro-average of the benchmark results using real notes for the source in RE-ID and DE-ID were 0.746 and 0.754, respectively, where the macro-average was 0.723 in RE-ID and 0.728 in DE-ID. Notably, on a micro average, RE-ID approaches across almost all settings were slightly better than DE-ID, with 6 out of 9. The Normalized One-shot generation also demonstrated the highest classification, and even GPT3.5 reached 0.749, exceeding the RE-ID benchmark. One-shot generation showed higher variation depending on the models. Particularly, Mistral7B reached around 0.74 in both RE-ID and DE-ID in the micro-average, which revealed comparable performance near to the benchmark. For the Keyword generation, the performance was around 0.71, with a larger variance across three models in the micro-average. On the contrary, the Normalized One-shot revealed the smallest variance in the 10-fold cross-validation.

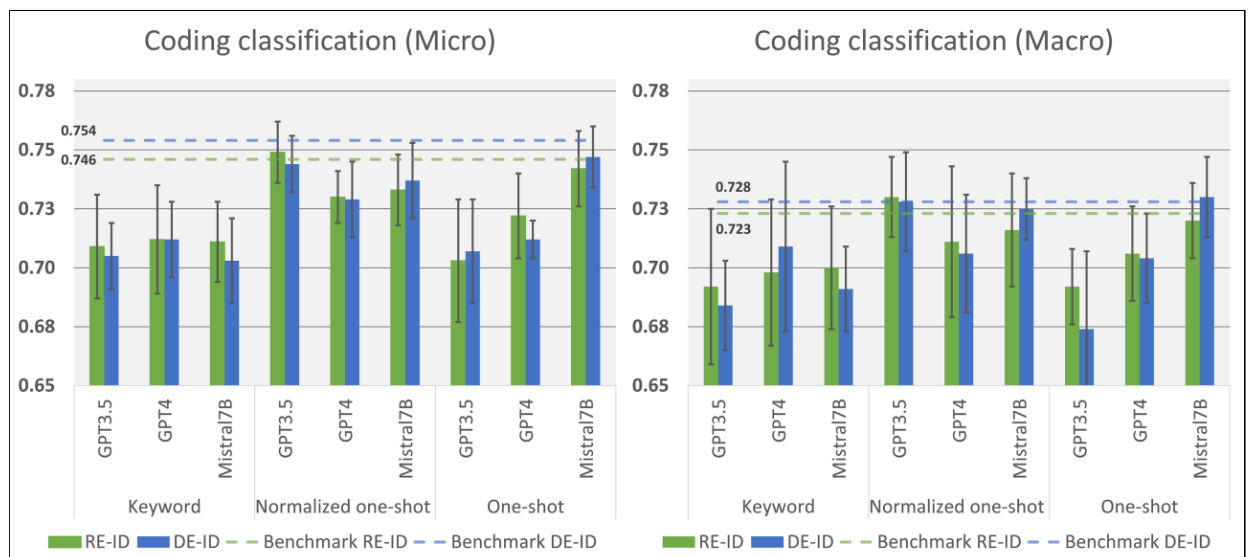

**Figure 2.** Bar chart of classification performance on ICD-9 coding.

The evaluation of text similarity using ROUGE metrics is depicted in Figure 3, while COS-SIM metrics are shown in Figure 4. The overall ROUGE scores for comparisons between source and synthetic notes followed the same pattern, regardless of whether DE-ID or RE-ID methods were used. Specifically, the Mistral7B model excelled in the Normalized One-shot generation, achieving the highest F1 scores for ROUGE1-3. The GPT4 model consistently displayed moderate performance across all metrics. In contrast, the GPT3.5 model demonstrated poor performance in both Keyword and Normalized One-shot generations, except the One-shot generation, where it achieved the highest scores across all ROUGE metrics.

When comparing different generation types, the Keyword generation method exhibited similar scores across all three models and displayed the lowest performance. In the Normalized One-shot generation, the ROUGE scores were the lowest for GPT-3.5, followed by GPT-4, with Mistral7B achieving the highest. In the One-shot generation, both GPT-3.5 and Mistral7B outperformed other generation types. The detailed values can be found in Table 5 of Supplementary Appendix 1.

COS-SIM scores were assessed for RE-ID and DE-ID methodologies, comparing source notes to synthetic notes. In the RE-ID setting, both One-shot and Keyword generations yielded similar COS-SIM scores across all three models, approximately 0.8 and 0.69, respectively. In the Normalized One-shot generation, the Mistral7B model achieved the highest cosine similarity, showing competitive results comparable to those in the One-shot approaches. Notably, the GPT3.5 model exhibited significantly poorer performance in the Normalized One-shot method, with COS-SIM scores around 0.55 for RE-ID and DE-ID settings. Moreover, both Keyword and One-shot generations in the RE-ID context slightly outperformed those in DE-ID. These findings highlight the diverse strengths of each generation type and model in maintaining text similarity in synthetic note generation.

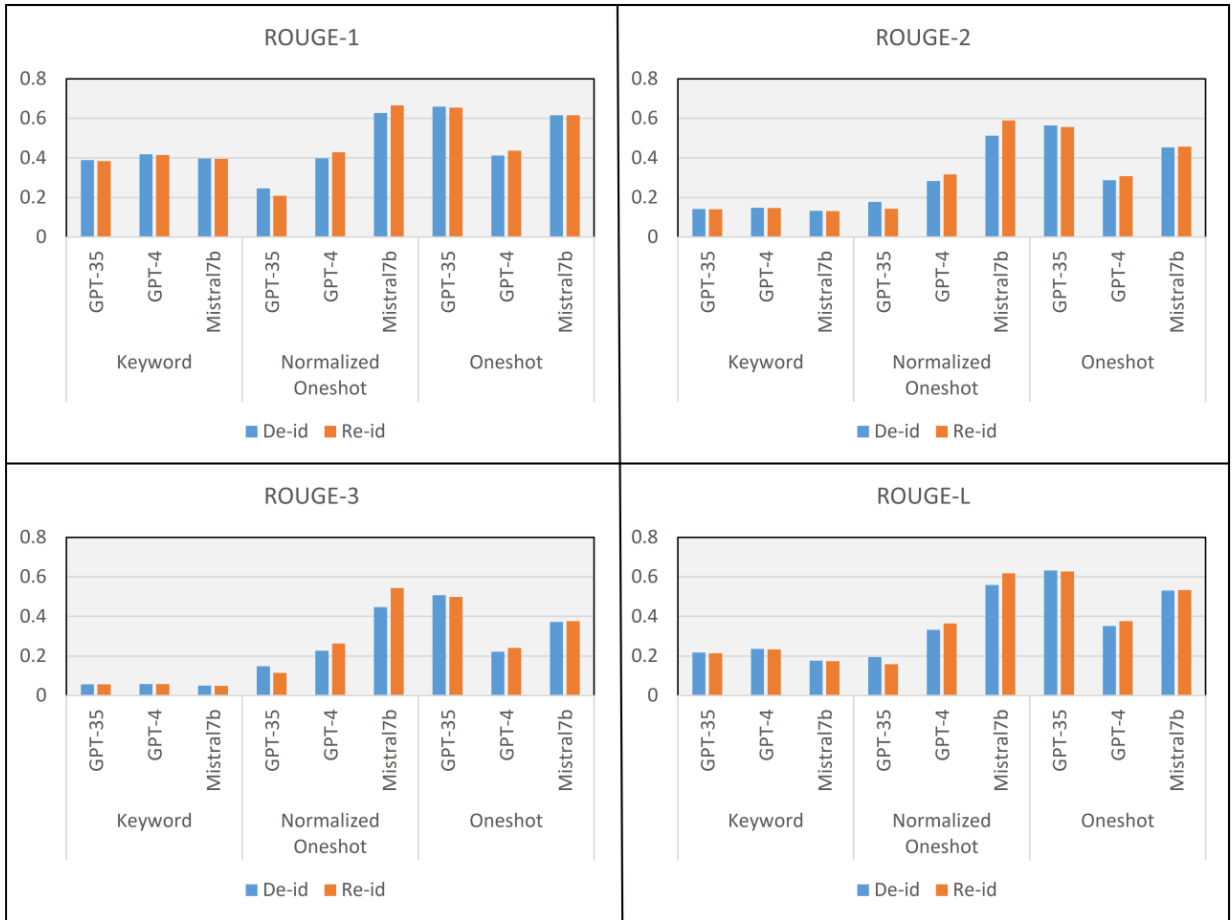

**Figure 3.** Bar charts comparing ROUGE scores between synthetic and original notes.

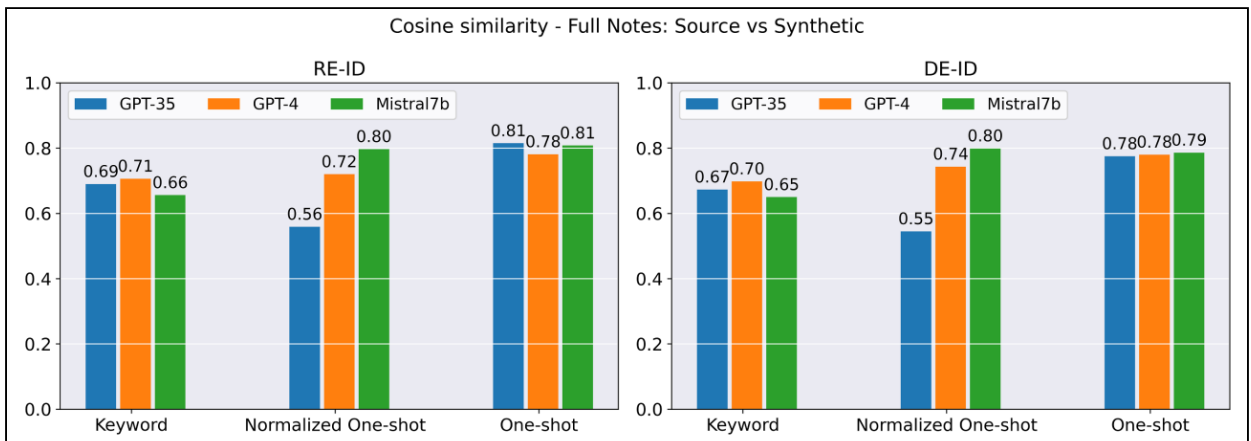

**Figure 4.** Bar charts comparing COS-SIM between synthetic and original notes.

## Discussion

*Text generation: Privacy and Utility*

The Normalized One-shot provided the best usability in synthetic note production, and the Keyword generation approach offered high anonymity. The only PHI identified was due to a geographic location present within the keyword prompts, which can be readily removed during pre-processing before the data is input into LLMs for text generation. Incorporating this feature into the pipeline allows privacy-free notes to be generated safely. Moreover, these secure notes achieved performance comparable to actual notes in ICD-9 coding classification by retaining meaningful tokens from the source notes. This method demonstrated excellent generalizability with current state-of-the-art LLMs. Although there was a performance trade-off in the Keyword generation for coding classification, adjustments to the TF-IDF threshold or incorporating additional features for keyword selection could enhance performance further. The decrease in ROUGE scores and COS-SIM in similarity assessments could be attributed to removing less critical elements and PHIs from the note, which helps preserve privacy.

Conversely, while popular and straightforward, the One-shot generation approach could significantly compromise patient privacy, particularly where high PHI co-occurrence might directly link to the original subject. In summary, synthetic notes generated via the keyword approach can effectively anonymize patient data while maintaining richness and usability, potentially transforming how clinical notes are shared and utilized. This approach holds significant potential to revolutionize the sharing and utilization of clinical notes.

*Re-identification: Enhancing Data Privacy in Clinical Data Usage*
Another significant finding from this study was that re-identified data exhibited greater usefulness and reliability compared to de-identified data in synthetic clinical notes. Regarding ICD-9 coding classification, most results showed that re-identified data through Keyword and Normalized One-shot generation outperformed de-identified data. Repopulating de-identified PHI tokens with dummy PHIs can enhance the ability of LLMs to generate more informative synthetic notes, thereby improving the utility of such synthetic notes for subsequent downstream tasks. This discovery constitutes a novel contribution to the field of data anonymization. It is noteworthy that conventional practices in institutions often prioritize data de-identification using placeholder or redaction tags to facilitate private data sharing. However, current de-identification methods have yet to meet the standards set by the HIPAA Safe Harbor method. Furthermore, the expert determination method is costly and time-consuming, presenting challenges in accurately assessing privacy risks. Building upon this insight, re-populating PHI placeholders in de-identified notes with dummy PHIs can safeguard patient privacy better while enhancing the utility of Keyword and Normalized One-shot text generation.

Additionally, in the era of generative AI, our comprehensive assessment of actual PHI has established an essential benchmark for privacy analysis in LLMs. Our analysis revealed that geographic terms are highly recognizable by LLMs and were easily reused in outputs. Additionally, unique identifiers such as unique identification numbers, medical record numbers, and device identifiers, rarely included in LLMs' general training data, could be wholly extracted and reappear in the outputs, thus posing a privacy risk. Consequently, these insights are crucial for the real-world application of these technologies in medical institutions, providing important references for managing privacy risks effectively.

*Limitations*
This study had several limitations that warrant mention. First, one limitation of this study is the inability to directly compare the proposed method with existing NER models, as our evaluation of PHI prevalence is both unique and novel. Current PHI de-identification models do not fully address the removal of all PHI categories required for HIPAA Safe Harbor compliance and often face challenges in balancing model generalizability with strict compliance standards[12–14]. As a result, existing models are not fully aligned with the compliance objectives addressed by our approach, making direct comparisons difficult. Second, the re-identification process might only partially represent the diversity of situations found in EHRs. For instance, specific settings such as holidays were designated simply as the word "Holiday" combined with a note's ID. This approach might not capture the nuanced variations that occur in real-world scenarios. In practical applications, the keyword process might require additional rules to ensure such terms are excluded. Third, the MIMIC III dataset used was preprocessed in prior work and was limited to single coding scenarios. While this earlier work provided a useful foundation, real-world applications of EHRs often involve complex, multi-coding scenarios. Future studies could benefit from a broader analysis that includes these complexities to enhance the findings. Forth, the error analysis on the performance of LLMs is limited, particularly when smaller LLMs outperform larger ones. This variability in performance is influenced by differences in pre-training data, model architecture, and task-specific optimizations[27]. These factors affect model effectiveness across tasks, underscoring the need for detailed and standardized evaluation to select the best-fitting LLMs[28]. Lastly, given the rapid development of technology and models, the field of EHRs and synthetic note generation is continuously evolving. Consequently, the findings of this study may become outdated as new methods and models are introduced. To provide a more comprehensive evaluation of the utility and effectiveness of synthetic notes, it is essential to explore alternative coding classifications and emerging technologies.

**Conclusion**

This study highlights the efficacy of the Keyword-based conditional generation approach in producing synthetic clinical notes that protect privacy while retaining usability. By anonymizing data yet maintaining the essence of the original notes, this method demonstrates a potent solution in how clinical data could be shared and utilized. Moreover, the comparative success of re-identified data over de-identified data suggests a shift towards methods that enhance both data utility and privacy. While the approach promises significant advancements, the need for further refinement and exploration of diverse clinical scenarios remains. Future research should continue to advance this promising field, focusing on improving methodologies in line with technological progress.

**Acknowledgments:**
We would like to express our sincere gratitude to all the authors for their invaluable support and guidance throughout this study. Their expertise and contributions have been essential to the success of this research.

**Funding**
A.R.S. is a Gordon P. Osler scholar and was partially supported by UMGF fellowship. N.M. was supported by the NSERC Discovery Grants (RGPIN-04127-2022), and Falconer Emerging Researcher Rh Award. X.J. is CPRIT Scholar in Cancer Research (RR180012). He was supported in part by the Christopher Sarofim Family Professorship, UT Stars award, UTHealth startup, the National Institute of Health (NIH) under award number R01AG066749, R01LM013712, R01LM014520, R01AG082721, U01AG079847, U24LM013755, U01TR002062, U01CA274576, and the National Science Foundation (NSF) #2124789.

**Conflicts of interest**
None declared.

**Pipeline Availability:**
The source code is available at https://github.com/lifestrugglee/Privacy-Synthetic-Generation. Specifically, the GPT models' API utilized in this study is the HIPAA-compliant Azure OpenAI platform provided by UTHealth to ensure compliance with the data usage agreement requirements of MIMIC-III.

**Supplementary Appendix 1**

Table 1: Details and Guidelines for Dataset Re-Identification

| NO. | MIMIC III PHI Categories | HIPAA Categories | Re-identified settings |
|---|---|---|---|
| 1 | Address | Geographic | Generated using the random-address package[26]. |
| 2 | Age* | All date | Random number from 52-99. |
| 3 | Clip number | Unique identifying number | A four-digit random number prefixed by "*clip_*" could be represented as "*clip_1234*". |
| 4 | Company | Geographic | A note ID prefixed by "*Company*". |
| 5 | Country | Geographic | Randomly selected from a list of 196 countries. |
| 6 | Date | All date | A year consisting of four digits. |
| 7 | First name | Names | Randomly selected from a list of 300 first names. |
| 8 | Full name | Names | Combined a *first name* and *last name*, resulting in 90,000 combinations. |
| 9 | Full name phone | Names, Telephone | Combined the procedure for *full name* and *phone number*. |
| 10 | Holiday | All date | A note ID prefixed by "*Holiday*". |
| 11 | Hospital | Geographic | A note ID prefixed by "*Hospital*". |
| 12 | Hospital ward | Geographic | A note ID prefixed by "*Hospital Ward*". |
| 13 | Job number | Unique identifying number | A five-digit random number prefixed by "*jn_*" could be represented as "*jn_12345*". |
| 14 | Last name | Names | Randomly selected from a list of 300 last names. |
| 15 | Location | Geographic | Utilized the city components when using the random-address package. |
| 16 | MD number | Unique identifying number | A combination of two characters with a four-digit random number prefixed by "*MD_*" can be represented as "*MD_XY_1234*". |
| 17 | Medical record number | Medical record numbers | A six-digit random number prefixed by "*mrn_*" could be represented as "*mrn_123456*". |
| 18 | Name initial | Names | Similar to the full name procedure, but retain only the first letter. |
| 19 | Number | Unique identifying number | A three-digit random number. |
| 20 | Numeric identifier | Unique identifying number | A six-digit random number prefixed by "*ni_*" could be represented as "*ni_123456*". |
| 21 | Pager number | Telephone | A random number between 10 and 999, prefixed by "*pg_*", could be represented as "*pg_123*". |
| 22 | Phone number | Telephone | Random numbers are represented in the format *123-45-6789*. |
| 23 | Serial number | Device identifiers | A seven-digit random number prefixed by "*sn_*" could be represented as "*sn_1234567*". |
| 24 | SSN | SSN | A random number formatted like a SSN (e.g. *123-45-6789*). |

| 25 | State | Geographic | The U.S. consists of 50 states, in addition to the District of Columbia, Puerto Rico, and the Virgin Islands. |
| 26 | Unit number | Unique identifying number | A four-digit random number prefixed by "*un_*" could be represented as "*un_1234*". |
| 27 | University | Geographic | A note ID prefixed by "*University_*". |

*Note:*
*Following the MIMIC III subject description, the median age of the adult patients is 65.8 years, with the interquartile range (IQR) extending from 52.8 years (Q1) to 77.8 years (Q3). In order to cover the age above 89, the final settings include a median age of 65.8 years, a first quartile (Q1) of 52 years, and a third quartile (Q3) of 99 years, defining a distribution centered around older adults with a wide spread, capturing variability in age with potential for outliers.

| | |
|---|---|
| Input | ```{<br>  "role": "user",<br>  "content": "Write a description of a patient using 2250 words containing the following phrases sequentially: *date discharge date date, service surgery allergies patient recorded, complaint abdominal pain major surgical, invasive procedure ercp, stent placement laparoscopic converted, open cholecystectomy history……*}``` |
| Output | On the date of admission, the patient presented with a complaint of abdominal pain. The patient had a history of major surgical procedures, including an invasive procedure called ERCP, which involved stent placement. The initial plan was for a laparoscopic cholecystectomy, but the procedure was converted to an open cholecystectomy due to complications. The patient had a …… |

**Figure 1.** Illustrative example of prompts and synthetic notes generated using the Keyword generation approach.

| | |
|---|---|
| Input | ```{<br>  "role": "user",<br>  "content": "As a physician, please write a clinical note using the following template.<br>           *TARGET TEMPLATE NOTE*"<br>}``` |
| Output | On the date of admission, a 24-year-old male patient was recorded to have a chief complaint of respiratory weakness. The patient had a history of myasthenia with bulbar symptoms and diplopia. He was subsequently found to have high achr titers and rep stim testing was done. The patient had undergone major surgical procedures in the past and had been intubated for plasmapheresis. |

**Figure 2.** Illustrative example of prompts and synthetic notes generated using the One-shot generation approach. The Normalized One-shot generation utilized the identical prompt as employed in the One-shot generation.

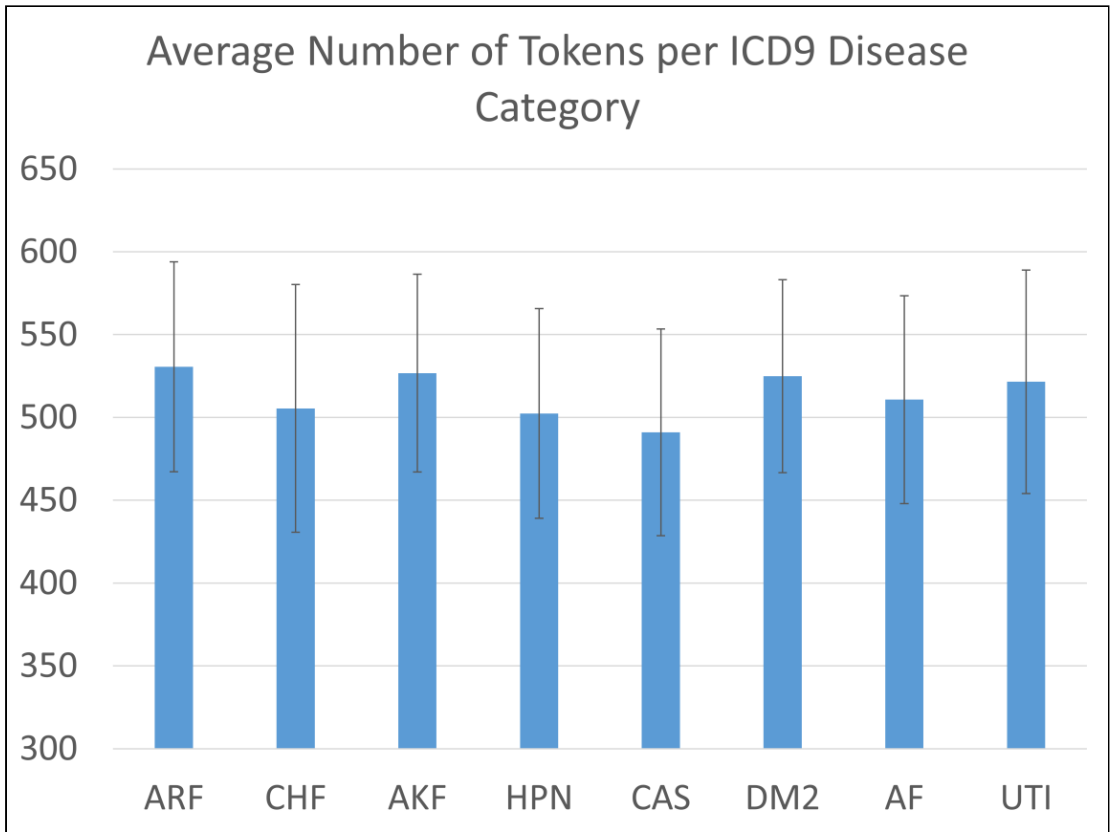

**Figure 3.** Mean number of extracted keyword tokens across ICD9 disease categories. ARF: Acute respiratory failure; CHF: Congestive heart failure; AKF: Acute kidney failure; HPN: Hypertension; CAS: Coronary atherosclerosis; DM2: Diabetes mellitus type 2; AF: Atrial fibrillation; UTI: Urinary tract infection

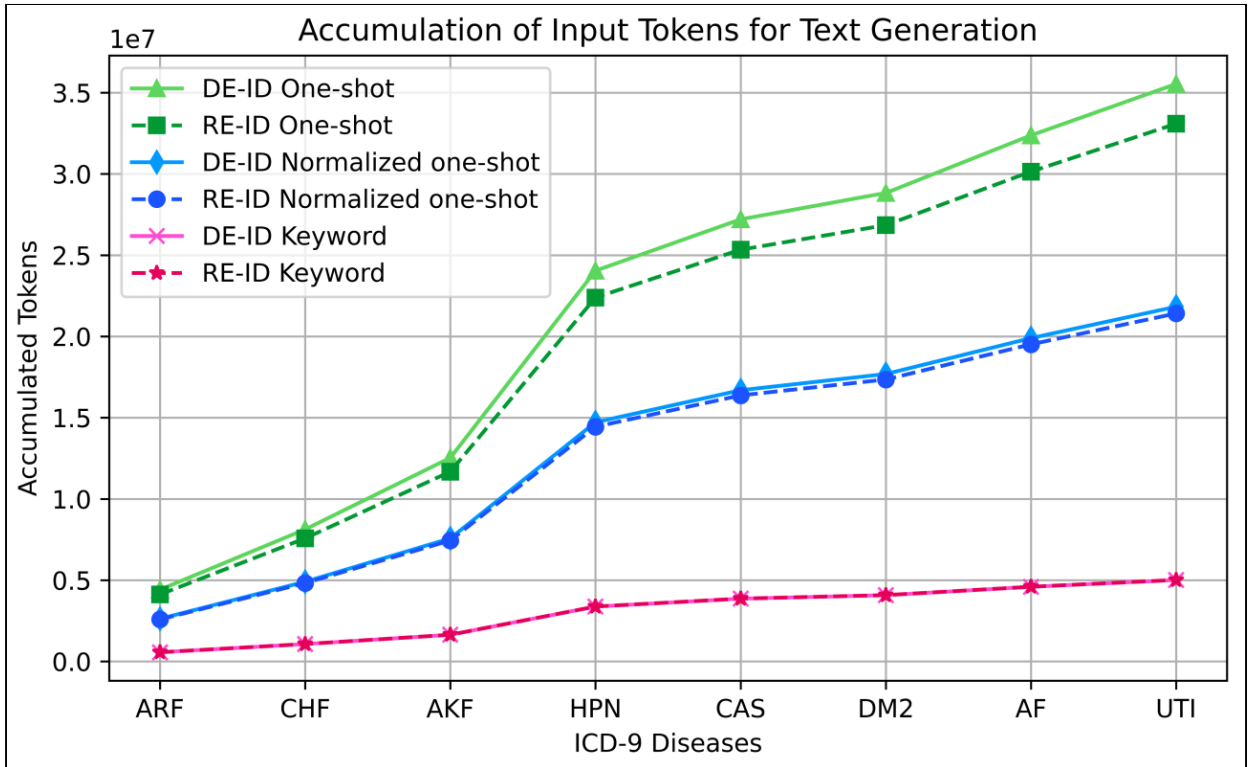

**Figure 4.** Accumulated input tokens for re-identified and de-identified data across various text generation methods. ARF: Acute respiratory failure; CHF: Congestive heart failure; AKF: Acute kidney failure; HPN: Hypertension; CAS: Coronary atherosclerosis; DM2: Diabetes mellitus type 2; AF: Atrial fibrillation; UTI: Urinary tract infection

Table 2: The total cost among each text generation approach.

| Models | Generation type | DE-ID | RE-ID |
|---|---|---|---|
| **GPT3.5** | One-shot | $ 51.36 | $ 52.00 |
| | Normalized one-shot | $ 78.77 | $ 75.35 |
| | Keyword | $ 26.30 | $ 26.17 |
| **GPT4** | One-shot | $ 1,260.84 | $ 1,155.81 |
| | Normalized one-shot | $ 1,671.99 | $ 1,505.97 |
| | Keyword | $ 734.60 | $ 727.61 |

*Note:* For GPT-3.5, the cost per thousand tokens is $0.002 for both input and output. In contrast, for GPT-4, the cost per thousand tokens is $0.03 for input and $0.06 for output.

| Table 3: ICD-9 coding classification performance on DE-ID data | | | |
|---|---|---|---|
| **Generation type** | **Models** | **Micro** | **Macro** |
| **Keyword** | GPT3.5 | 0.705 ± 0.014 | 0.684 ± 0.019 |
| | GPT4 | 0.712 ± 0.016 | 0.709 ± 0.036 |
| | Mistral7B | 0.703 ± 0.018 | 0.691 ± 0.018 |
| **Normalized one-shot** | GPT3.5 | 0.744 ± 0.012 | 0.728 ± 0.021 |
| | GPT4 | 0.729 ± 0.016 | 0.706 ± 0.025 |
| | Mistral7B | 0.737 ± 0.016 | 0.725 ± 0.013 |
| **One-shot** | GPT3.5 | 0.707 ± 0.022 | 0.674 ± 0.033 |
| | GPT4 | 0.712 ± 0.008 | 0.704 ± 0.019 |
| | Mistral7B | 0.747 ± 0.013 | 0.730 ± 0.017 |

| Table 4: ICD-9 coding classification performance on RE-ID data | | | |
|---|---|---|---|
| **Generation type** | **Models** | **Micro** | **Macro** |
| **Keyword** | GPT3.5 | 0.709 ± 0.022 | 0.692 ± 0.033 |
| | GPT4 | 0.712 ± 0.023 | 0.698 ± 0.031 |
| | Mistral7B | 0.711 ± 0.017 | 0.700 ± 0.026 |
| **Normalized one-shot** | GPT3.5 | 0.749 ± 0.013 | 0.730 ± 0.017 |
| | GPT4 | 0.730 ± 0.011 | 0.711 ± 0.032 |
| | Mistral7B | 0.733 ± 0.015 | 0.716 ± 0.024 |
| **One-shot** | GPT3.5 | 0.703 ± 0.026 | 0.692 ± 0.016 |
| | GPT4 | 0.722 ± 0.018 | 0.706 ± 0.020 |
| | Mistral7B | 0.742 ± 0.016 | 0.720 ± 0.016 |

Table 5: ROUGE scores between synthetic and original notes across DE-ID and RE-ID

| | ROUGE-1 | | | | | | | | |
|---|---|---|---|---|---|---|---|---|---|
| | Keyword | | | Normalized One-shot | | | One-shot | | |
| Data type | GPT-35 | GPT-4 | Mistral7b | GPT-35 | GPT-4 | Mistral7b | GPT-35 | GPT-4 | Mistral7b |
| DE-ID | 0.388 | 0.418 | 0.397 | 0.246 | 0.398 | 0.627 | 0.660 | 0.412 | 0.615 |
| RE-ID | 0.384 | 0.415 | 0.395 | 0.208 | 0.429 | 0.666 | 0.654 | 0.437 | 0.615 |

| | ROUGE-2 | | | | | | | | |
|---|---|---|---|---|---|---|---|---|---|
| | Keyword | | | Normalized One-shot | | | One-shot | | |
| Data type | GPT-35 | GPT-4 | Mistral7b | GPT-35 | GPT-4 | Mistral7b | GPT-35 | GPT-4 | Mistral7b |
| DE-ID | 0.142 | 0.148 | 0.133 | 0.178 | 0.283 | 0.513 | 0.564 | 0.286 | 0.453 |
| RE-ID | 0.140 | 0.147 | 0.132 | 0.143 | 0.317 | 0.590 | 0.557 | 0.308 | 0.457 |

| | ROUGE-3 | | | | | | | | |
|---|---|---|---|---|---|---|---|---|---|
| | Keyword | | | Normalized One-shot | | | One-shot | | |
| Data type | GPT-35 | GPT-4 | Mistral7b | GPT-35 | GPT-4 | Mistral7b | GPT-35 | GPT-4 | Mistral7b |
| DE-ID | 0.057 | 0.058 | 0.050 | 0.148 | 0.227 | 0.446 | 0.507 | 0.221 | 0.372 |
| RE-ID | 0.056 | 0.058 | 0.049 | 0.114 | 0.263 | 0.544 | 0.498 | 0.240 | 0.375 |

| | ROUGE-L | | | | | | | | |
|---|---|---|---|---|---|---|---|---|---|
| | Keyword | | | Normalized One-shot | | | One-shot | | |
| Data type | GPT-35 | GPT-4 | Mistral7b | GPT-35 | GPT-4 | Mistral7b | GPT-35 | GPT-4 | Mistral7b |
| DE-ID | 0.217 | 0.235 | 0.176 | 0.195 | 0.332 | 0.559 | 0.632 | 0.351 | 0.530 |
| RE-ID | 0.214 | 0.233 | 0.174 | 0.158 | 0.364 | 0.618 | 0.627 | 0.376 | 0.534 |

*Note:* A darker color indicates a higher value.